\documentclass[11pt,a4paper]{article}
\usepackage[nohyperref]{naaclhlt2019}
\usepackage{times}
\usepackage{latexsym}
\usepackage{adjustbox}
\usepackage{url}
\usepackage{footnote}
\usepackage{multirow}
\makesavenoteenv{table}
\makesavenoteenv{tabular}

\aclfinalcopy 

\title{Generate, Filter, and Rank: Grammaticality Classification for Production-Ready NLG Systems}

\author{
  Ashwini Challa\thanks{\ \ Equal contribution} \\
  \And
  Kartikeya Upasani$^{*}$\\
  \And
  Anusha Balakrishnan \\
  \And
  Rajen Subba\\
  \AND 
  Facebook Conversational AI\\
  \texttt{\{ashwinichalla, kart, anushabala, rasubba\}@fb.com}
}

\date{}

\begin{document}
\maketitle
\begin{abstract}
  Neural approaches to Natural Language Generation (NLG) have been promising for goal-oriented dialogue. One of the challenges of productionizing these approaches, however, is the ability to control response quality, and ensure that generated responses are acceptable. We propose the use of a generate, filter, and rank framework, in which candidate responses are first filtered to eliminate unacceptable responses, and then ranked to select the best response. While acceptability includes grammatical correctness and semantic correctness, we focus only on grammaticality classification in this paper, and show that existing datasets for grammatical error correction don't correctly capture the distribution of errors that data-driven generators are likely to make. We release a grammatical classification and semantic correctness classification dataset for the weather domain that consists of responses generated by 3 data-driven NLG systems. We then explore two supervised learning approaches (CNNs and GBDTs) for classifying grammaticality. Our experiments show that grammaticality classification is very sensitive to the distribution of errors in the data, and that these distributions vary significantly with both the source of the response as well as the domain. We show that it's possible to achieve high precision with reasonable recall on our dataset.
\end{abstract}

\section{Introduction}
In recent years, neural network-based approaches have been increasingly promising in the context of goal-oriented Natural Language Generation (NLG). These approaches can effectively learn to generate responses of desired complexity and detail from unaligned data. Additionally, these approaches can be scaled with relatively low effort to new domains and use cases. However, they are less robust to mistakes and have poor worst case performance. Consistently achieving acceptable response quality in a customer facing product is an immediate blocker to using such models widely.

Controlling quality at generation time in these models is challenging, and there are no guarantees that any of the generated responses are suitable to surface to an end user. Additionally, quality is hard to enforce at data collection time, given the increasingly widespread dependence on large pools of untrained annotators. As a result, classifying acceptability with high precision is extremely desirable. It can be used to establish safe fallbacks to acceptable, but potentially less ideal, responses that are generated by more traditional NLG systems like templates. Such responses are likely to be grammatically and semantically correct, but may sacrifice detail, variety, and naturalness; this trade-off may sometimes be necessary in a consumer-facing product. For example, the system could respond with ``Here's your weather forecast'', and show a card with relevant weather information, rather than generate an incoherent weather forecast.

Some key aspects of acceptability are \textbf{grammaticality} and \textbf{semantic correctness}. A grammatical response is one that is well-formed, and a semantically correct response is one that correctly expresses the information that needs to be conveyed. Systems that generate ungrammatical or incorrect responses run the risk of seeming unreliable or unintelligent. Another important facet of acceptability is the \textbf{naturalnesss} (or human likeness) of the response, that can improve the usability of chatbots and other dialogue systems.

In this paper, we first propose the inclusion of a filtering step that performs acceptability classification in the more widely used generate \& rank framework (Generate, Filter, and Rank). Then, we narrow our focus to grammaticality classification, and show how this problem calls for datasets of a different nature than typical grammatical error correction (GEC) datasets. We also show that state-of-the-art GEC models trained on general corpora fail to generalize to this problem. Finally, we introduce a dataset of system-generated grammatical errors for the \textbf{weather} domain, and demonstrate the performance of some strong baselines for grammatical classification on this data.  This dataset can also be used for further research on semantic correctness classification. Our experiments also reinforce the need for the new framework we propose.

\section{Generate, Filter, and Rank}
\label{sec:genfilterrank}
In this section, we first review the pros and cons of the traditional generate \& rank framework, and then propose a ``filter'' step that addresses some of its downsides.

The generate \& rank framework has been proposed and widely used in several prior works on goal-oriented dialogue (\citet{Walker2001spot}, \citet{Langkilde1998}). In NLG systems, the typical use of this framework involves generating multiple candidate responses (often using various different surface realization techniques), and then reranking these using statistical models (most commonly language models). More recent works have also proposed reranking to optimize for certain personality traits or user engagement (\citet{soundingboard}). The input to the generators is usually a structured representation of what the system needs to convey. 

This setup allows for the use of multiple generator models, as proposed in \citet{milabot} and \citet{alquist}, among others. This greatly increases the number of possible responses that can be surfaced, which can improve both diversity and naturalness. The use of statistical rerankers also allows systems under this framework to optimize for naturalness as well as acceptability (primarily grammaticality), since typical statistical models should easily be able to downrank potentially ungrammatical candidates.  However, there are a few practical concerns that arise with using this framework in production:
\begin{enumerate}
    \item Data sparsity: The space of unseen named entities like locations, datetimes, etc., and other sparse token types is potentially very large. This can result in suboptimal language modeling behaviors, in which language models downrank valid candidates with sparse surface forms.
    \item Statistical models that are typically used for reranking cannot capture semantic correctness without conditioning on the goal and arguments. They also run the risk of accidentally biasing towards more likely (but semantically incorrect) responses. This is particularly tricky for ML-based generators, where the generated responses can easily leave out important information. For example, the best models from \citet{nayak2017} have error rates between 2-5\%.
    \item There is a significant risk that none of the responses generated by data-driven models is acceptable. For example, in the dataset that we release in this work, there were no grammatical responses generated for around 12\% of the scenarios (see Section \ref{sec:dataset}).
\end{enumerate}

The common thread in these issues is that the generate \& rank framework conflates acceptability, which is objective, with naturalness and other traits, which are subjective. To address, we propose the addition of a high-precision ``filter" step that eliminates any unacceptable responses before the ranking stage, allowing the reranker to focus on optimizing for naturalness and other desirable properties. Since we found grammaticalness to be a more serious issue than semantic correctness in our dataset (Table \ref{table:basic_stats}), we explore methods to implement a grammaticality ``filter'' in the following sections.

\begin{table*}
\centering
\resizebox{0.95\textwidth}{!}{
\begin{tabular}{|p{0.45\linewidth}|p{0.95\linewidth}|}
\hline
\textbf{Error Category} & \textbf{Examples} \\
\hline
Repeated words like ``with", ``and". & In Grand Prairie , it 's 100 degrees fahrenheit \textbf{with} cloudy skies \textbf{with} snow showers. \\
\hline
Agreement & Friday, September 15 in Branford , it'll be cloudy with a high of 73 degrees fahrenheit with \textbf{an 61} percent chance of snow showers . \\
\hline
Dangling modifiers & In Tongan Qu on Monday, May 22 \textbf{will be} scattered clouds with Patches of Fog , with a high of 18 degrees celsius and a low of 7 degrees . \\
\hline
Incorrect word choice & In Larne on Thursday, November 23 , \textbf{it'll be scattered clouds} with Fog , with a high of 46 and a low of 35 degrees fahrenheit. \\
\hline
Ungrammatical n-grams & In Funabashi-shi on Monday, March 20 , there will be a low of 31 with a high of 47 degrees fahrenheit with \textbf{scattered clouds skies} and a Light Drizzle \\
\hline
Missing contextual words, like ``degrees" & In Caloocan City , expect a temperature of \textbf{3 celsius} with mostly sunny skies and Fog Patches \\
\hline
Linking words/phrases & Right now in Arrondissement de Besancon , it 's 2 degrees fahrenheit \textbf{with sunny and Light Fog} \\
\hline
\end{tabular}
}
\caption{Mistakes involving grammatical errors and other cases of unacceptability in model-generated weather responses}
\label{table:error_distribution}
\end{table*}

\vspace{0.2cm}

\resizebox{0.45\textwidth}{!}{
\begin{tabular}{|p{0.5\linewidth}|p{0.30\linewidth}|p{0.30\linewidth}|}
\hline
    \hfill & \textbf{CoNLL-2014} & \textbf{Our dataset}\\
\hline
\# grammatical & 53426 & 18494 \\
\hline
\# ungrammatical & 21638 & 14511 \\
\hline
\% scenarios with no grammatical responses & N/A & 12\% \\
\hline
Avg. length & 22.8 & 17.9 \\
\hline
Vocab size & 28180 & 5669\\
\hline 
\# goals & N/A & 2\\
\hline
\# semantically correct & N/A & 28475\\
\hline
\# semantically incorrect & N/A & 4530\\
\hline
\end{tabular}
}
\captionof{table}{Comparison of weather responses dataset against the NUCLE corpus} 
\label{table:basic_stats}

\begin{table*}[t]
\centering
\resizebox{\textwidth}{!}{
\begin{tabular}{|c|c|c|c|c|c|c|}
\hline
\multicolumn{1}{|c|}{}
&
\multicolumn{2}{|c|}{Train} 
&
\multicolumn{2}{|c|}{Eval} 
&
\multicolumn{2}{|c|}{Test} \\
\hline
  \textbf{Generator} & \textbf{\#grammaticals} & \textbf{\#ungrammaticals} & \textbf{\#grammaticals} & \textbf{\#ungrammaticals} & \textbf{\#grammaticals} & \textbf{\#ungrammaticals}\\
\hline
SC-LSTM Lex & 4957 & 2386 & 1565 & 882 & 1712 & 757 \\
\hline
SC-LSTM Delex & 1083 & 2078 & 365 & 679 & 377 & 657 \\
\hline
IR & 1530 & 2513 & 532 & 839 & 493 & 833 \\
\hline
Gen LSTM & 3614 & 1624 & 1133 & 600 & 1247 & 549 \\
\hline
\end{tabular}
}
\caption{Distribution of positive and negative examples in weather responses dataset} 
\label{table:per_model_dist}
\end{table*}

\section{Mismatched Error Distributions }
The CoNLL-2014 shared task on grammatical error correction (\citet{ng2014}) released the NUCLE corpus for grammatical error correction (GEC), written by students learning English. Ungrammatical sentences in this dataset contain annotations and corrections of each individual error. From a classification perspective, each original ungrammatical utterance in the dataset is a negative example, and the final corrected utterance (obtained by applying all of the corrections to the original ungrammatical utterance) is a positive example. Additionally, sentences without any corrections are positive examples as well. 

These positive and negative samples can then be directly used to train the grammaticality filter described in previous sections. In the runtime of the goal-oriented NLG system, this filter would be used to filter out ungrammatical responses that are generated by \textbf{models} - even though the filter was trained on human-written responses. This signals the possibility of a data mismatch. 

To better understand the nature of this difference, we collected a corpus of system-generated responses for the \textbf{weather} domain (see Section \ref{sec:dataset}) and manually inspected \textasciitilde200 of these responses to identify common categories of model mistakes (see Table \ref{table:error_distribution}). Interestingly, we found that the most common mistakes made by our models, like repeated words and missing contextual words, don't match any of the error categories in NUCLE (see Table 1 from \citet{ng2014}). There are also qualitative differences stemming from the domains in these datasets. Our corpus has a large number of mentions of sparse entities (particularly locations), dates, and weather-specific constructs like temperatures, while the NUCLE corpus is open-ended and spans a variety of topics.

In order to quantify this difference, we measure the performance of open-domain GEC models on our corpus by evaluating a model that achieves state-of-the-art performance on the CoNLL-2014 test set \citep{gecmlconv}. We found that this model failed to generalize well to our dataset (see section \ref{sec:expts}), and missed several classes of errors. For example, the model failed to catch any of the errors in Table \ref{table:error_distribution} (see Appendix \ref{sec:appendix} for more examples).

Intuitively, this suggests that training models for response filtering demands datasets very different in distribution from publicly available datasets that only reflect human mistakes. We show this empirically through experiments in section \ref{sec:expts}, and describe the process for collecting our dataset in the next section.

\section{Dataset}
\label{sec:dataset}
We first collected a dataset of human-generated responses for the weather domain, using a process similar to the one used in \citet{E2EDataset}. Each of the collected responses is conditioned on a \textbf{scenario}, consisting of a goal (the intent to be expressed) and arguments (information to be expressed). In collecting the dataset, we restricted ourselves to the goals \texttt{inform\_current\_condition} and \texttt{inform\_forecast}.

An example scenario is \\ ``requested\_location": ``London", \\ ``temp": ``32", \\ ``temp\_scale": ``fahrenheit", \\ ``precip\_summary": ``Heavy Blowing Snow" \\
A possible response for this scenario is \texttt{In London, it's currently 32 degrees Fahrenheit with heavy snow.}.

We then trained some standard NLG models on this corpus. Two of these (\texttt{sc-LSTM Lex} and \texttt{sc-LSTM Delex}) are semantically conditioned LSTMs as described in \citet{SCLSTM}; the \texttt{genLSTM} model is a vanilla LSTM decoder; and \texttt{IR} is a simple retrieval-based generator. The details of these are described in Appendix \ref{sec:modeldetails}. We generated $n=3$ responses from each of these models for each scenario in a held out data set, and deduped generated candidates that differed by a single character (often punctuation). We then asked crowdworkers to judge the grammaticality of these responses. Our final dataset\footnote{\texttt{github.com/facebookresearch/momi}} consists of 33K model-generated responses with grammaticality and semantic correctness judgments. Table \ref{table:per_model_dist} shows a detailed breakdown of grammatical and ungrammatical responses per model.


\section{Approach}\label{sec:approach}
\textbf{Preprocessing}
We made the assumption that the specific values of arguments such as locations, dates, and numbers do not affect sentence framing. We therefore replaced locations and dates with placeholder tokens. Numbers are replaced with either \textit{\_\_num\_\_}, \textit{\_\_num\_vowel\_\_} if the number begins with a vowel sound (example, 80), or \textit{\_\_num\_one\_\_} if the number is 1. Hence the sentence \textit{``There is an 85 percent chance of rain in New York on Wednesday, August 25"} would become \textit{``There is an \_\_num\_vowel\_\_ percent chance of rain in \_\_location\_\_ on \_\_date\_\_"}. In case of \texttt{sc-LSTM delex}, all remaining arguments (such as weather conditions) are also delexicalized.

To maintain class balance in the train set, for each response source, the class with fewer samples is upsampled to match the number of samples of the other class. When training on samples from multiple generators, the samples of each generator in train set are upsampled to match those of generator with highest number of samples. Upsampling is not done for validation or test sets.

\textbf{Gradient Boosted Decision Tree Using LM Features (LM-GBDT)}\label{sec:GBDT}
Language models (\citet{ngramlm}) can effectively capture n-gram patterns that occur in grammatical data, making features derived from them good candidates to distinguish between grammatical and ungrammatical responses. We train a 7-gram LM\footnote{We found that 7 gram LM performed slightly better than other lower n-gram LMs. LM with larger n-grams may be better at catching model mistakes that require looking at long-range dependencies.} on human-written weather responses described in Section \ref{sec:dataset}. The trained LM is then used to extract features listed in Table \ref{table:LMFeatures} for each model-generated response. Finally, we feed these features into a gradient boosted decision tree (GBDT) (\citet{gbdt}) to classify the model-generated response as grammatical or not.

\vspace{0.2cm}

\resizebox{0.45\textwidth}{!}{
\begin{tabular}{|p{0.5\linewidth}|p{0.5\linewidth}|}
\hline
\multicolumn{2}{|c|}{\textbf{Features from Language Model}} \\
\hline
Geometric mean: $\left(\Pi_{i=1}^{m}p_i\right)^{(1/m)}$  & Arithmetic mean: $\sum_{i=1}^{m}p_i / m$\\
\hline
$\min P_x$ & $\max P_x$\\
\hline
Median: $\tilde p$ & Std Dev: $\sigma_{P_x}$\\
\hline
\multicolumn{2}{|c|}{$C_{P_{x}}(0, 0.1)$, $C_{P_{x}}(0.1, 0.2)$, .... $C_{P_{x}}(0.9, 1.0)$}\\
\hline
\end{tabular}
}
\captionof{table}{Features derived from Language Model. $P_x = {p_1, p_2, ..... p_m}$ is the set of all n-gram probabilities from an n-gram LM for a sentence $x$. $C_{P_{x}}(a,b) \in [0, 1]$ is the ratio of n-gram probabilities $p_i \in P_x$ for which $a \leq p_i < b$.} \label{table:LMFeatures}

\begin{figure}
  \caption{CNN-based grammaticality classifier}
  \label{figure:docnn}
  \centering
  \includegraphics[width=0.48\textwidth]{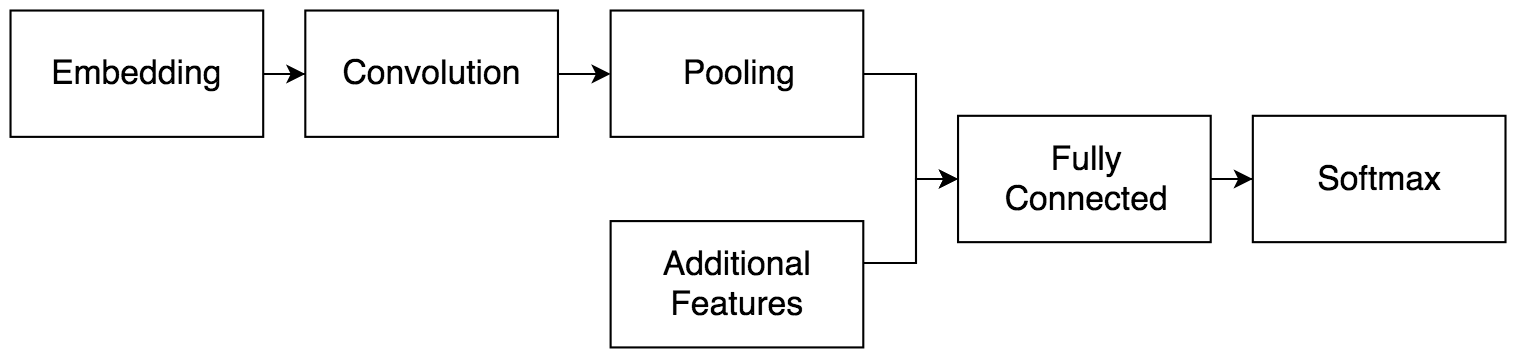}
\end{figure}

\textbf{CNN-based Classification Model}
We used a convolutional neural network (CNN) for sentence classification in an approach similar to \citet{kim2014docnn}. Figure \ref{figure:docnn} illustrates the architecture of the model. After pooling convolutional features along the time dimension, the result can be optionally concatenated with additional features. A one-hot vector of length 4 encoding the source of the response (IR, \texttt{GenLSTM}, \texttt{sc-LSTM delex}, \texttt{sc-LSTM lex}) is passed as an additional feature when training on responses from multiple sources.

\section{Experiments}\label{sec:expts}
We try different combinations of NUCLE corpus and our dataset as train and test sets to learn a grammaticality classifier for model-generated weather responses. Table \ref{table:results6.2} and \ref{table:results6.3} lists the results of these experiments described above. As discussed before, since the goal is to build a classifier for use in production systems, we report the recall of models for grammatical class when the precision is very high (98\%). In cases where the model does not achieve this precision, we report recall at the highest precision achieved by the model. 

\texttt{CNN + source} represents the case when the source of response is passed as an additional feature to the CNN. We used filters with widths 2, 3, 4, 5 for the CNN. Performance did not change with different number and sizes of filters.

\vspace{0.2cm}
 
\resizebox{0.48\textwidth}{!}{
\begin{tabular}{|p{0.3\linewidth}|p{0.4\linewidth}|p{0.5\linewidth}|}
\hline
Experiment setting & Ungrammatical response picked & Semantically incorrect response picked\\
\hline
Ranker          & 29.4\% & 8.2\%  \\
\hline
Filter + Ranker & 2.4\%  & 0.75\% \\
\hline
\end{tabular}
}
\captionof{table}{Comparison of number of times the top ranked response is unacceptable with and without filtering.} \label{table:rankerVsFilter}

\vspace{0.2cm}

\resizebox{0.48\textwidth}{!}{
\begin{tabular}{|p{0.3\linewidth}|p{0.4\linewidth}|p{0.2\linewidth}|p{0.2\linewidth}|p{0.25\linewidth}|}
\hline
  \textbf{Model} & \textbf{Training Data} & \textbf{Test Data} & \textbf{R@P98} & \textbf{R@P}   \\
\hline
\citet{gecmlconv} & NUCLE & Weather & - & 75 @ 64 \\
\hline
\multirow{3}{*}{CNN}& \multirow{2}{*}{NUCLE} & NUCLE  & 62.4 & -\\
\cline{3-5}
 & & Weather & - & 80 @ 56.8 \\
\cline{2-5}
 & NUCLE + Weather & Weather & 52.5 & -  \\
\cline{1-5}
 CNN	& Weather & Weather & 71.9 & - \\
\cline{1-5}
 CNN + source & \textbf{Weather} & \textbf{Weather} & \textbf{72.8} & - \\
\cline{1-5}
 LM-GBDT & Weather & Weather & 63.8 & - \\
\hline
\end{tabular}
}
\caption{Training on NUCLE and weather data} 
\label{table:results6.2}

\vspace{0.2cm}

\resizebox{0.48\textwidth}{!}{
\begin{tabular}{|p{0.17\linewidth}|p{0.4\linewidth}|p{0.4\linewidth}|p{0.2\linewidth}|p{0.23\linewidth}|}
\hline
  \textbf{Model} & \textbf{Training Data} & \textbf{Test Data} & \textbf{R@P98} & \textbf{R@P}   \\
\hline
\multirow{8}{*}{\begin{minipage}{0.5in}CNN + source\end{minipage}} & Weather & \texttt{IR} & 9.8 & - \\
\cline{2-5}
	& \texttt{IR} & \texttt{IR} & 23.2 & - \\
\cline{2-5}
	& Weather & \texttt{GenLSTM} & 95.5 & - \\
\cline{2-5}
	& \texttt{GenLSTM} & \texttt{GenLSTM} & 92.2 & - \\
\cline{2-5}
	& Weather & \texttt{SC-LSTM Delex} & 25.2 & - \\
\cline{2-5}
	& \texttt{SC-LSTM Delex} & \texttt{SC-LSTM Delex} & - & 45.9@80 \\
\cline{2-5}
 & Weather & \texttt{SC-LSTM Lex} & 96.8 & - \\
\cline{2-5}
 & \texttt{SC-LSTM Lex} & \texttt{SC-LSTM Lex} & 94.6 & - \\
\cline{1-5}
\multirow{8}{*}{\begin{minipage}{0.5in}LM- GBDT\end{minipage}} & Weather & \texttt{IR} & - & 8@95.5 \\
\cline{2-5}
 & \texttt{IR} & \texttt{IR} & 18 & - \\
\cline{2-5}
 & Weather &\texttt{GenLSTM} & 83.4 & - \\
\cline{2-5}
 & \texttt{GenLSTM} &\texttt{GenLSTM} & 76 & - \\
\cline{2-5}
 & Weather & \texttt{SC-LSTM Delex} & 2 & -  \\
\cline{2-5}
 & \texttt{SC-LSTM Delex} & \texttt{SC-LSTM Delex} & - & 65.5@70.5  \\
\cline{2-5}
 & Weather & \texttt{SC-LSTM Lex} & 90.6 & - \\
\cline{2-5}
 & \texttt{SC-LSTM Lex} & \texttt{SC-LSTM Lex} & 88.4 & - \\
\hline
\end{tabular}
}
\caption{Performance of filter for individual generators} 
\label{table:results6.3}

\subsection{Ranker vs Filter + Ranker}\label{subsec:rankervsfilter}
In order to validate the Generate, Filter, and Rank framework, we used our trained n-gram language model\footnote{an n-gram based language model is a simple baseline. It is possible to use more sophisticated rankers (such as RNN-LMs) to achieve better results. However, ranking approaches will still fail over filters when there are no grammatical candidates at all.} (from Section \ref{sec:GBDT}) to rank all the responses for each scenario in our dataset. We then measured the \% of times the top ranked candidate is ungrammatical, to understand how many times the final response would be ungrammatical in a traditional generate \& rank framework. We repeat the experiment with our proposed framework, by filtering ungrammatical responses using a CNN-based filter with 98\% precision before the ranking step. The results are shown in Table \ref{table:rankerVsFilter}.

The filtering step increases the overall response quality, but comes at the cost of losing genuine grammatical candidates because of slightly lower recall, 72.8\%, (the best recall we achieved on the weather data set). This is a fundamental tradeoff of our proposed framework; we sacrifice recall for the sake of precision, in order to ensure that users of the system very rarely see an unacceptable response. The semantic correctness also improves, but this doesn't indicate that grammatical filter is enough to solve both grammaticalness and correctness problems.

\subsection{Performance of filters on NUCLE and weather data}
Table \ref{table:results6.2} compares performance of CNN, LM-GBDT, and the GEC model used by \citet{gecmlconv}. The GEC model is adopted for binary classification by checking whether the model makes a correction for an ungrammatical sentence, and doesn't make any corrections for a grammatical sentence \footnote{We assume that the GEC model has classified the response as ungrammatical if an edit is made. This does not account for cases in which the edited response is still ungrammatical. As a result, the precision of this model in the true setting would be lower than that reported in this setting.}. This model achieves poor precision and recall on our dataset, and we found that it fails to generalize adequately to the novel error types in our data.
 
We also train the CNN on NUCLE data and find that it similarly achieves poor recall when classifying weather responses. This is attributed to the fact that the domain and category of errors in both datasets are different. Comparing Table 1 in \citet{ng2014} and Table \ref{table:error_distribution} of this work further supports this observation.

The CNN and LM-GBDT are trained and tested on our weather dataset. We report the performance of these models on the \textbf{complete} weather test set, not just on individual generators, since this is closest to the setting in which such models would be used in a production system. The CNN consistently has better recall than LM-GBDT at the same precision. \texttt{CNN + source} performs better than the CNN, indicating that information regarding source helps in classifying responses from multiple generators.

Augmenting the weather responses with NUCLE corpus while training the CNN did not help performance.

\subsection{Performance of filter for individual generators}
Table \ref{table:results6.3} presents results on test sets of each generator for classifiers trained together on all generators and trained on individual generators. Models trained individually on \texttt{IR} and \texttt{SC-LSTM Delex} responses perform poorly compared to \texttt{GenLSTM} and \texttt{SC-LSTM Lex} as the training set size is much smaller for former. The recall for individual generators is higher when training is done on data from all generators, indicating that the approach generalizes across sources. An exception to this is \texttt{IR} which does better when trained just on \texttt{IR} responses. This may be due to errors of retrieval based approach being different in nature compared to LSTM-based approach.

Table \ref{table:error_examples_appendix} in Appendix shows the errors in responses from different generators. Some errors occur more frequently with one generator than another, for example, the problem of repeating words (like \textit{with} and \textit{and}) is dominant in responses generated by the LSTMs, but very rarely seen in \texttt{IR} since it is a retrieval based approach.

\subsection{Comparison of LM-GBDT and CNN}
The recall of CNN is slightly better than LM-GBDT consistently across experiments. Both approaches do well in catching types of errors listed in Table \ref{table:error_distribution}. One difference between the two is the ability of CNN-based models to successfully catch errors such as ``1 degrees", while the LM-GBDT fails to do so. On further inspection, we noticed that the human generated weather responses, which were used as training data for the language model, contained several instances of ``1 degrees". The LM-GBDT has a heavy dependency on the quality of features generated by LM (which in turn depends on the quality of the LM training corpus), and this is a disadvantage compared to the CNNs.

\section{Related Work \& Conclusion}
Several previous works have established the need for a generate \& rank framework in a goal-oriented NLG system (\citet{Walker2001spot}, \citet{Langkilde1998}). Recent work on the Alexa prize (\citet{alexaprize}) has demonstrated that this architecture is beneficial for systems that bridge the gap between task-oriented and open-ended dialogue (\citet{milabot}, \citet{soundingboard}, \citet{alquist}). In such systems, the ranker needs to choose between a much more diverse set of candidates, and potentially optimize for other objectives like personality or user satisfaction. To make such systems practical for production-scale usage, our work proposes the inclusion of a high precision filter step that precedes ranking and can mark responses as acceptable. Our experiments show that this filter with sufficient fallbacks guarantees response quality with high precision, while simply reranking does not (Section \ref{subsec:rankervsfilter}).

In this work, we focus specifically on filtering ungrammatical responses. Previous work in this space has focused on classifying (and sometimes correcting) errors made by humans (\citet{ng2014}) or synthetically induced errors (\citet{foster2007treebanks}). We found, however, that the domain and error distribution in such datasets is significantly different from that of typical data-driven generation techniques. To address this gap, we release grammatical and semantic correctness classification data generated by these models, and present a reasonable baseline for grammatical classification. The approaches we present are similar to work on grammatical classification using features from generative models of language, like language models (\citet{gramClassification}). One future direction is to explore modeling semantic correctness classification with the datatset we release.

We compare the performance of two approaches for classifying grammaticality: CNNs, and GBDTs with language model features. Both are standard classifiers that are easy to deploy in production systems with low latency. An interesting future direction would be to explore model architectures that scale better to new domains and generation approaches. This could include models that take advantage of existing GEC data consisting of human responses, as well as datasets similar to ours for other domains. Models that successfully make use of these datasets may have a more holistic understanding of grammar and thus be domain- and generator-agnostic.

A drawback of the generate-filter-rank framework is the increased reliance on a fallback response in case no candidate clears the filtering stage. This is an acceptable trade-off when the goal is to serve responses in production systems where the standards of acceptability are high. One way to alleviate this is to do grammatical error correction instead of simply removing unacceptable candidates from the pipeline. Correcting errors instead of rejecting candidates can be of value for trivial mistakes such as missing articles or punctuation. However, doing this with high precision and correcting semantic errors remains a challenge.



\bibliography{naaclhlt2019}
\bibliographystyle{acl_natbib}

\appendix

\section{Appendices}
\label{sec:appendix}
\section{NLG Models for Generating Weather Responses}\label{sec:modeldetails}
The dataset we present in this paper consists of responses generated by 4 model types:
\begin{enumerate}
    \item \texttt{sc-LSTM delex}: An sc-LSTM trained on fully delexicalized human responses, where delexicalization refers to the process of replacing spans corresponding to specific arguments by placeholder strings.
    \item \texttt{sc-LSTM lex}: An sc-LSTM trained on partly delexicalized human responses. For this model, we only delexicalize locations, dates, and temperatures, thus allowing the model to freely choose surface forms for any other arguments.
    \item \texttt{GenLSTM}: A vanilla LSTM-based decoder model, where the decoder hidden state is initialized using embeddings of the goal and arguments. This model is also trained on fully delexicalized responses.
    \item \texttt{IR}: A simple retrieval approach in which \texttt{n} random candidates that satisfy the given goal and arguments are retrieved. The retrieved candidates are delexicalized, and any candidates that contain the right arguments (regardless of argument value) are considered valid.
\end{enumerate}
For all models, the final response is obtained by replacing argument placeholders by the canonical values of those arguments in the scenario.

Since our goal was just to get responses from a diverse set of data-driven generators with a reasonable distribution of errors, we did not experiment too much with improving \texttt{IR} and \texttt{genLSTM}, which are much weaker than the sc-LSTM models.

\section{Model-Generated Responses: Error Analysis}\label{appendixa}

Table \ref{table:error_examples_appendix} shows errors made by different generators. While there is an overlap in the category of grammatical errors made by different generators, the frequency of the errors is largely different. There are also a few generator-specific errors. For example, the problem of repeating words (like \textit{with} and \textit{and}) is dominant in responses generated by \texttt{GenLSTM}, \texttt{sc-LSTM delex}, \texttt{sc-LSTM lex}, but very rarely seen in IR. This is because human responses themselves are unlikely to have repeating words, however the LSTM-based generators tend to make these mistakes while trying to fit all information into the response. Ungrammatical n-grams like \textit{scattered clouds skies} are very infrequent in \texttt{sc-LSTM lex} responses while more commonly seen with other generators. This is because the \texttt{sc-LSTM lex} generator directly produces surface forms of weather conditions. LSTM models doesn't tend to generate responses with out of vocabulary words, but it is something common with IR responses usually because of spelling mistakes in templates.

\begin{center}
\begin{table*}
\resizebox{\textwidth}{!}{
\centering
\begin{tabular}{|p{0.2\linewidth}|p{0.25\linewidth}|p{0.8\linewidth}|}
\hline
 \textbf{Generator} & \textbf{Error Category} & \textbf{Examples} \\
\hline
\multirow{6}{*}{SC-LSTM Lex} & Repeating words like ``with", ``and". & In Grand Prairie , it 's 100 degrees fahrenheit \textbf{with} cloudy skies \textbf{with} snow showers. \\
\cline{2-3}
& Poor choice of words to connect 2 phrases & Right now in Medford , \textbf{with} a temperature of -10 degrees celsius .\\
\cline{2-3}
& Wrong Plurals/singulars & In Yushu , it's \textbf{1 degrees} celsius and cloudy . \\
\cline{2-3}
& Missing words that forms incomplete sentences & In Tongan Qu on Monday, May 22 \textbf{will be} scattered clouds with Patches of Fog , with a high of 18 degrees celsius and a low of 7 degrees .\newline \newline * ... May 22 there will be scattered clouds ... * \newline \newline Right now in East Liverpool it is \textbf{-3 fahrenheit} with Heavy Rain.\newline \newline *missing word degrees* \\
\cline{2-3}
& Wrong articles & Friday, September 15 in Branford , it'll be cloudy with a high of 73 degrees fahrenheit with \textbf{an 61} percent chance of snow showers . \\
\cline{2-3}
& Wrong pronouns & In Larne on Thursday, November 23 , \textbf{it'll be scattered clouds} with Fog , with a high of 46 and a low of 35 degrees fahrenheit. \newline \newline *there'll be scattered clouds* \\
\hline
\multirow{5}{*}{SC-LSTM Delex} & Repeating words like ``with", ``and" & In Chengtangcun on Wednesday, April 12 expect a high of 2 degrees and a low of -10 degrees celsius \textbf{with} cloudy skies \textbf{with} Snow Showers . \\
\cline{2-3}
& Wrong word choices & In Funabashi-shi on Monday, March 20 , there will be a low of 31 with a high of 47 degrees fahrenheit with \textbf{scattered clouds skies} and a Light Drizzle \\
& Wrong articles & In Newbury on Tuesday, February 07 , there will be \textbf{an 46} percent chance of Heavy Rain Showers with a high of 5 degrees celsius with overcast skies . \\
\cline{2-3}
& Wrong Pluarals/Singulars & In Shiselweni District on Tuesday, March 21 , it will be overcast with a high of 8 degrees celsius and a low of \textbf{1 degrees} . \\
\cline{2-3}
& Missing contextual words like ``degrees" & In Caloocan City , expect a temperature of \textbf{3 celsius} with mostly sunny skies and Fog Patches \\
\hline
\multirow{6}{*}{Gen LSTM} & Repeating words like ``with", ``and" & Right now in Wojewodztwo Maopolskie , it 's sunny \textbf{with} Light Thunderstorms \textbf{with} Hail and a temperature of 13 degrees fahrenheit . \\
\cline{2-3}
& Poor word choices & Right now in Franklin Square , it 's 96 degrees fahrenheit with \textbf{scattered clouds skies} . \\
\cline{2-3}
& Wrong articles & On Friday, November 17 in San-Pedro , expect a low of 44 and a high of 68 degrees fahrenheit with \textbf{an 41} percent chance of Flurries .\\
\cline{2-3}
& Wrong Plurals/Singulars & Right now in Minnetonka Mills , it 's \textbf{1 degrees} celsius with sunny skies .\\
\cline{2-3}
& Wrong pronouns & On Monday, July 03 in Himeji Shi , \textbf{it'll be scattered clouds} with a high of 48 degrees fahrenheit.\newline \newline*there'll be scattered clouds* \\
\cline{2-3}
& Wrong connecting words & Right now in Arrondissement de Besancon , it 's 2 degrees fahrenheit \textbf{with sunny and Light Fog}\newline \newline*... and sunny with light fog ... would make it grammatical* \\
\hline
\multirow{6}{*}{IR} & Wrong articles & In Shiraki Marz on Thursday, November 09, there will be \textbf{an 51} percent chance of Heavy Blowing Snow and a high of 39 degrees fahrenheit \\
\cline{2-3}
& Wrong ordinal indicators & On Friday, June \textbf{02th} in Selma there will be a low of 82 degrees fahrenheit with Light Thunderstorms with Hail \\
\cline{2-3}
& Wrong Plurals/Singulars & On Tuesday, June 13, in Wilayat-e Paktiya, there will be Heavy Snow Showers and the high will be \textbf{1 degrees} celsius. \\
\cline{2-3}
& Wrong helping verbs (Plural versus singular) & N/A \\
\cline{2-3}
& Wrong Pronoun & On Wednesday, October 18, in Reus, \textbf{it'll be scattered clouds} and 3 degrees celsius.\newline \newline *... there'll be scattered clouds ...* \\
\cline{2-3}
& Poor templates like one with repeating words, spelling mistakes, missing words like degrees & In Rudraprayag on Tuesday, November Tuesday, June 13 temp is \textbf{-8 to 0 celsius} with Low Drifting \textbf{Snow Snow} Showers and overcast cloud \\
\cline{2-3}
& Out of vocabulary words & It's currently -15 degrees fahrenheit \textbf{t} and mostly clear with gentle breeze in Dammam \\
\hline
\end{tabular}
}
\caption{Some more examples of grammatical errors made by different generation models in our dataset.} \label{table:error_examples_appendix}
\end{table*}
\end{center}

\section{General GEC Model Performance}
Table \ref{table:gec_error_examples_appendix} shows examples of ungrammatical responses that the general GEC model (\citep{gecmlconv}) failed to correct, and Table \ref{table:gec_corrected_examples_appendix} shows examples of ungrammatical responses that the model correctly edited. The model corrects mistakes that are much more likely to occur in the GEC data (like verb agreement), but fails to catch model-specific error types like stuttering and other ungrammatical n-grams.

\begin{table*}
\centering
\resizebox{0.95\textwidth}{!}{
\begin{tabular}{|p{0.75\linewidth}|p{0.3\linewidth}|}
\hline
\textbf{Response} & \textbf{Error type}\\
\hline
On Friday, February 17 in Changwat Samut Songkhram, expect \textbf{a likely of heavy rain showers} and a high of 15 degrees celsius. & Agreement\\
\hline
Currently in Maastricht there is fog and is -3 Fahrenheit. & Missing ``degrees''\\
\hline
Right now in Westminster it is \textbf{1 degrees} Fahrenheit with partly cloudy skies. & Numerical agreement \\
\hline
In Ayacucho on Wednesday, February 22,\textbf{ with a} high of 83 degrees fahrenheit \textbf{with a} 98 percent chance of light snow. & Repeated ``with'', and missing linking words \\
\hline 
In kajiado on Thursday, January 12, expect a high of 82 degrees and a low of 61 degrees Fahrenheit \textbf{with mostly sunny}. & Incomplete response \\
\hline

\end{tabular}
}
\caption{Mistakes involving grammatical errors and other cases of unacceptability in model-generated weather responses}
\label{table:gec_error_examples_appendix}
\end{table*}

\begin{table*}
\centering
\resizebox{0.95\textwidth}{!}{
\begin{tabular}{|p{0.5\linewidth}|p{0.5\linewidth}|}
\hline
\textbf{Original Response} & \textbf{Corrected Response}\\
\hline
The weather for Wednesday, December 27 in Oak Hill will includes a high of 14 Celsius and a 37 percent chance of heavy freezing rain. & The weather for Wednesday, December 27 in Oak Hill \textbf{will include} a high of 14 Celsius and a 37 percent chance of heavy freezing rain.\\
\hline
In Ocean County, it is 34 degrees Fahrenheit with sunny. & In Ocean County, it is 34 degrees Fahrenheit with sunny \textbf{weather}.\\
\hline
In Bim Son, it is 1 degrees fahrenheit with funnel cloud. & In Bim Son, it is 1 degrees fahrenheit with \textbf{funnel clouds}. \\
\hline

\end{tabular}
}
\caption{Mistakes involving grammatical errors and other cases of unacceptability in model-generated weather responses}
\label{table:gec_corrected_examples_appendix}
\end{table*}

\end{document}